\documentclass[twocolumn,times,final]{elsarticle}

\usepackage{prletters}
\usepackage{framed,multirow}

\usepackage{amssymb}
\usepackage{latexsym}

\usepackage{url}
\usepackage{xcolor}
\definecolor{newcolor}{rgb}{.8,.349,.1}
\usepackage{amsmath,amsfonts}
\usepackage{mathtools}
\usepackage{booktabs}
\usepackage{siunitx}
\usepackage{algorithmic}
\usepackage{algorithm}
\usepackage{array}
\usepackage[caption=false,font=normalsize,labelfont=sf,textfont=sf]{subfig}
\usepackage{footmisc}
\usepackage{textcomp}
\usepackage{stfloats}
\usepackage{url}
\usepackage{verbatim}
\usepackage{graphicx}
\usepackage{lineno}

\journal{Pattern Recognition Letters}

\usepackage{blindtext}
\usepackage{hyperref}
\usepackage{nameref}

\newcounter{mylabelcounter}

\makeatletter
\newcommand{\labelText}[2]{%
\refstepcounter{mylabelcounter}%
\immediate\write\@auxout{%
 \string\newlabel{#2}{{\unexpanded{#1}}{\thepage}{{\unexpanded{#1}}}{mylabelcounter.\number\value{mylabelcounter}}{}}%
}%
}
\makeatother

\begin{document}

\begin{frontmatter}

\title{Optimal word order for non-causal text generation
with Large Language Models: the Spanish case}

\author[1]{Andrea \surname{Busto-Castiñeira}\corref{cor1}} 
\cortext[cor1]{Corresponding author: abusto@gti.uvigo.es}
\ead{abusto@gti.uvigo.es}
\author[1]{Silvia \surname{García-Méndez}} 
\ead{sgarcia@gti.uvigo.es}
\author[1]{Francisco \surname{de Arriba-Pérez}} 
\ead{farriba@gti.uvigo.es}
\author[1]{Francisco J. \surname{González-Castaño}}
\ead{javier@det.uvigo.es}

\affiliation[1]{organization={Information Technologies Group, atlanTTic, University of Vigo},
        addressline={Rúa Maxwell, 16}, 
        city={Vigo}, 
        postcode={36310}, ,
        country={Spain}}

\begin{abstract}
Natural Language Generation (\textsc{nlg}) popularity has increased owing to the progress in Large Language Models (\textsc{llm}s), with zero-shot inference capabilities. However, most neural systems utilize decoder-only causal (unidirectional) transformer models, which are effective for English but may reduce the richness of languages with less strict word order, subject omission, or different relative clause attachment preferences. This is the first work that analytically addresses optimal text generation order for non-causal language models. We present a novel Viterbi algorithm-based methodology for maximum likelihood word order estimation. 
We analyze the non-causal most-likelihood order probability for \textsc{nlg} in Spanish and, then, the probability of generating the same phrases with Spanish causal \textsc{nlg}. This comparative analysis reveals that causal \textsc{nlg} prefers English-like \textsc{svo} structures. We also analyze the relationship between optimal generation order and causal left-to-right generation order using Spearman's rank correlation. Our results demonstrate that the ideal order predicted by the maximum likelihood estimator is not closely related to the causal order and may be influenced by the syntactic structure of the target sentence.\\

\end{abstract}


\end{frontmatter}

\section{Introduction}
\label{sec:introduction}
Thanks to the impressive recent advances in transformer-based Large Language Models (\textsc{llm}s) \citep{attention-2017, Webb2023}, nowadays, \textsc{llm}s are extensively used in diverse applications such as voice assistants \citep{Schobel2023}, chatbots \citep{Suhaili2021, Tiwari2023}, storytelling \citep{Nichols2021, Vartinen2022}, marketing \citep{Syrdal2023, Sood2023}, and even text-to-image models 
\citep{Gal2022, Youssef2023}.

Even though these data-intensive language models can write and interpret natural language with near-human fluency \citep{Floridi2020, Elkins2020}, there are still open challenges. For example, \textsc{llm}s can create biased and harmful output since the enormous training data sets are often scrapped from dubious sources \citep{Bender2021, Lucy2021}. The prevalence of English in most massive data sets is also a relevant bias source by itself \citep{Garrido-Munoz2021}. 
Since tokenizers assign longer tokens to character sequences found in languages with more representation \citep{Wu2020a, Rust2021}, linguistic data imbalance hurts the word embeddings of polyglot language models \citep{Liu2020}. Even if this particular issue has already been addressed in the literature \citep{Artetxe2019, Clark2022}, this is still pending.

Currently, the prevailing generative language models are decoder-only transformers, with OpenAI's GPT-4\footnote{Available at: \texttt{\url{https://cdn.openai.com/papers/gpt-4.pdf}}, October 2024} and ChatGPT\footnote{Available at: \texttt{\url{https://openai.com/blog/chatgpt}}, October 2024} being the most popular. Despite the theoretical multilingual capabilities of current OpenAI models, the performance of \textsc{llm}s is uneven across different languages \citep{Limisiewicz2023}.

One of the causes of this issue could be the underlying causality of state-of-the-art \textsc{nlg} \textsc{llm}s that are decoder-only transformers. They are expectation-based word predictors, producing text from left to right by recursively feeding themselves with previously generated sequences. Other expectation-based word predictors, such as Recurrent Neural Networks (\textsc{rnn}), have been shown to prefer temporal proximity between linked components within a sequence, which is also consistent with English grammar \citep{Davis2020}.

Expectation-based word prediction is heavily related to the psycholinguistic concept of surprise. Surprisal theory quantifies difficulty in sentence comprehension \citep{Hale2001, Levy2008, Henderson2016, Lowder2018}.
Even if generally accepted, surprisal theory does not model working memory in text comprehension, ignoring processing challenges in integrating words or components that are far apart within a text \citep{Gibson1998, Lewis2005, Bartek2011, Nicenboim2015, Nicenboim2016}. Lossy context surprisal \citep{Futrell2020} integrates expectation and memory-based predictability theories by representing working memory limitations as noise. Even though this model's premise is language-independent, it can accurately represent several language-specific text-processing phenomena.

Lossy context surprisal recreates structural forgetting by dropping part of the context and wrongly re-sampling it from the a priori language knowledge probability model. Structural forgetting \citep{Vasishth2010} is a common grammatical illusion in English in which ungrammatical double-embedded relative clauses are perceived as correct and vice-versa. Intrinsic probabilistic language expectations exclusively determine this, as both native and non-native speakers show structural forgetting in English. However, neither show it when presented with the same syntactic structures in German or Dutch \citep{Frank2016}. Even if prone to semantic illusions, Spanish is another representative case of a language that does not present structural forgetting \citep{Stroud2012}. This propensity of English probabilistic distribution to such backward prediction mistakes is coherent with the lack of success in English non-causal text generation.

Therefore, some languages may be more suited for non-causal language modeling since they do not exhibit structural forgetting and can be adequately predicted backward. Furthermore, structural forgetting implies that human text comprehension is not unidirectional, bringing the non-causal language models theoretically closer to human textual cognition.

In this work, we analyze the case of Spanish, which not only shows a preference for high nominal attachment in relative clause resolution and no structural forgetting but also has a highly flexible syntax \citep{Lahousse2012}, unlike the strict \textsc{s-v-o} structure of English grammar, with very few inversion exceptions \citep{Assaiqeli2021}.

We propose a technique based on the Viterbi algorithm \citep{Forney1973} to study the generation order in Spanish \textsc{nlg}. Using a non-causal language model, our approach computes the maximum-likelihood generation order for any Spanish sentence. Once the non-causal optimal generation order is determined, we compare the optimal order generation probability of a given set of sentences with that of a causal language model. The goal is to evaluate whether the use of causal language models for Spanish text generation restricts the syntactic richness of the produced text and to propose strategies to enhance the efficiency of Spanish \textsc{nlg} using non-causal language models. Even if our experiments focus on Spanish, the methodology proposed in this work applies to any other language by modifying the data set and language model used.

The Viterbi-based technique proposed is not a new natural language generation methodology but an \textit{a posteriori} optimal word order estimation method for non-causal language models. Even if not a generative approach by itself, our contribution will hopefully open the path towards more efficient and effective \textsc{nlg} algorithms for Spanish, a language with more than half a billion speakers, the fourth most spoken language in the world and second in terms of native speakers.

Summing up, the main contributions of this work are:
\begin{itemize}
  \item The first approach for estimating optimal word order for natural language generation.
  \item A novel comparative analysis to assess how the grammatical structure of a target sentence impacts the optimal generation order.
\end{itemize}

The rest of this article is organized as follows. Section \ref{sec:related} discusses related studies on non-causal \textsc{nlg} and other applications of the Viterbi algorithm in Natural Language Processing (\textsc{nlp}). Section \ref{sec:method} outlines the analytical framework used in this paper, presents our implementation of the Viterbi algorithm for maximum likelihood word generation order estimation, and presents our evaluation methodology. Section \ref{sec:results} describes the data set, details the model in our experimental validation, and discusses the obtained results thoroughly. Finally, Section \ref{sec:conclusions} concludes the paper, summarizing the main points and suggesting possible areas for further research.

\section{Related Works}
\label{sec:related}

To our knowledge, no previous works are analyzing non-causal \textsc{nlg} for non-English languages. This section will discuss relevant contributions to this work, mainly focusing on non-causal \textsc{nlg} and other Viterbi algorithm applications in \textsc{nlp}.

\subsection{Causality in generative transformer language models}
\label{sec:noncausal_related}

Transformers are unsupervised learners thanks to their self-attention mechanism \citep{attention-2017}, which controls the impact of the context on the model's output. The original transformer architecture is composed of an encoder and a decoder. While the encoder's attention is bidirectional, the decoder has a masked multi-head attention block that masks non-causal context and a bidirectional multi-head attention block that receives non-causal information from the encoder.

Although the encoder-decoder architecture is widely used in some \textsc{nlp} applications like machine translation \citep{Kawara2021, Nguyen2021}, other transformer-based models only use one of these two components. By excluding the encoder, we eliminate all non-causal contextual dependencies, thus using only the masked attention of the decoder. Currently, decoder-only transformers are the most effective task-agnostic \textsc{nlg} systems.

While open domain \textsc{nlg} is mainly causal, there are a few non-causal \textsc{nlg} solutions. Most non-causal \textsc{nlg} systems are focused on particular tasks such as speech recognition \citep{Chen2021, Wang2022}, style transfer and grammar correction \citep{Kaneko2020}, textual data augmentation \citep{Park2019}, and dialog systems \citep{Yu2021, Wang2023}.

Non-causal language models can also be trained for masked Language Modeling (\textsc{mlm}) \citep{Zeng2021}. \textsc{mlm} is an \textsc{nlg} task consisting of predicting masked words within a sentence. Some generative systems use bidirectional transformers trained on this task to recursively generate and fill masked tokens \citep{Shen2020}. As these can be filled in any location within the text, these models can produce text in a non-causal way.

Non-causal \textsc{nlg} strategies perform much worse in English than their causal counterparts \citep{Wang2019a}. However, to our knowledge, no prior research has been conducted on non-causal \textsc{nlg} in languages other than English. This work aims to evaluate whether bidirectional transformers trained on the \textsc{mlm} task could be successfully exploited in Spanish \textsc{nlg}.

\subsection{Viterbi algorithm in \textsc{nlp} }
The Viterbi algorithm \citep{Forney1973} is a dynamic programming algorithm used to find the most probable sequence of hidden states, known as the Viterbi path, given a sequence of observed events.

The Viterbi algorithm has been used in \textsc{nlp} applications such as speech recognition \citep{Jo2019,Raj2022}, in which the audio data is the observed sequence of events and the state space is defined by tokenized text strings, and in part-of-speech (\textsc{pos}) tagging \citep{Pattnaik2020}, in which the observed sequence is the given text and the state space is given by all possible \textsc{pos} tags. 

Even conceptually close, the Viterbi algorithm formulation proposed in this paper is different from that of those other \textsc{nlp} works. In our approach, instead of the tokenizer's vocabulary or \textsc{pos} tags, the Markov chain states are given by the initial phrase in all the possible different states of completion. The transition probabilities are computed using the bidirectional transformer language model described in Section \ref{sec:model_results}.

As previously said, our Viterbi algorithm-based approach explained in Section \ref{sec:method} is, to our knowledge, the first to estimate the optimal order for word generation in a language, contributing to the development of new non-causal \textsc{nlg} methodologies.

\section{Methodology}
\label{sec:method}
\subsection{Language modeling and sentence probability}

Language models can provide a token $\hat{X}$ probability distribution based on a context token sequence $X$. For a generation index $i<N$, we define the $N$-length input causal context as follows:

\begin{equation}
  \mathbf{x}_{\textsc{c}_i} = \begin{bmatrix}x_{i-N} & \dots & x_{i-1} \end{bmatrix}^T
\end{equation}

And the non-causal context of length $N$ as:

\begin{equation}
\mathbf{x}_{\textsc{nc}_i} = \begin{bmatrix} x_0 & \dots & x_{i-1} & x_{i+1} & \dots & x_{N} \end{bmatrix}^T
\end{equation}

We denote the output of a causal language model (\textsc{c-lm}) as:

\begin{equation}  \mathrm{\textsc{c-lm}}_{i}\left(\mathbf{x}_{\textsc{c}_i}\right) = \begin{bmatrix} P(\hat{X}_i = v_0 \mid X_{\textsc{c}} = \mathbf{x}_{\textsc{c}_i}) \\ \vdots \\ P(\hat{X}_i = v_{\left|\mathcal{V}\right|-1} \mid X_{\textsc{c}} = \mathbf{x}_{\textsc{c}_i} ) \end{bmatrix}
\end{equation}

While for a non-causal language model (\textsc{nc-lm}), the output is:

\begin{equation}
\mathrm{\textsc{nc-lm}}_{i}\left(\mathbf{x}_{\textsc{nc}_i}\right) = \begin{bmatrix} P(\hat{X}_i = v_0 \mid X_{\textsc{nc}} = \mathbf{x}_{\textsc{nc}_i}) \\ \vdots \\ P(\hat{X}_i = v_{\left|\mathcal{V}\right|-1} \mid X_{\textsc{nc}} = \mathbf{x}_{\textsc{nc}_i} ) \end{bmatrix}
\end{equation}

For, in both cases, a token vocabulary set $\mathcal{V} = \left\{\begin{matrix} v_0 & \dots & v_{\left|\mathcal{V}\right|-1}\end{matrix}\right\}$ and a vocabulary size $\left|\mathcal{V}\right|$.

The output probability mass distribution provided by these language models is sampled for text generation while recursively feeding the previously generated sequences as context.

In this work, we want to obtain the probability $\mathcal{P}_{\textsc{lm}}\left(\mathbf{t} \mid \mathcal{X}\right)$ with which a language model \textsc{lm} would generate a given token sequence $\mathbf{t} = \begin{bmatrix}
  t_0 & t_1 & \dots & t_N
\end{bmatrix}^T$ with a generation order $\mathcal{X}$:

\begin{equation}
  \mathcal{P}_{\textsc{lm}}\left(\mathbf{t} \mid \mathcal{X}\right) = \displaystyle \prod_{k=0}^{N-1} P\left(\hat{X}_k = t_k \mid X = \mathbf{x}_{k}, \textsc{lm}\right) = \prod_{k=0}^{N-1} \mathrm{\textsc{lm}}_{t_k}\left(\mathbf{x}_k\right)
\end{equation}

With $\mathbf{x}_k$ being the available context to generate token $t_k$. Note how $\mathbf{x}_k$ values vary with different generation orders $\mathcal{X}$. For the causal \textsc{nlg} case, $\mathbf{x}_k = \begin{bmatrix}
  t_0 & t_1 & \dots & t_{k-1}
\end{bmatrix}^T$ for all $k$, but it may not be the case for non-causal \textsc{nlg}.

\subsection{Viterbi maximum likelihood generation order estimation}
We will model our generation order estimation problem as a hidden Markov model (\textsc{hmm}) in which we predict the hidden generation order state sequence $\mathcal{X}$ from a state set $\mathcal{S}$ including all possible masked token combinations from our observation sequence $\mathbf{t}$. 
Using a language model \textsc{lm}, we can use an implementation of the Viterbi algorithm to compute the maximum likelihood generation order such that:

\begin{equation}
\mathcal{X}_{\mathrm{ML}} = \operatorname*{argmax}_\mathcal{X} \mathcal{P}_{\textsc{lm}}\left(\mathbf{t} \mid \mathcal{X}\right)
\end{equation}
Having $|\mathcal{S}| = \sum_{i=0}^{N} \binom{i}{N}$ Markov chain states and $|\mathcal{T}| = \sum_{i=0}^{N} \binom{i}{N}\cdot i$ transitions, the cost $c_k$ associated with a transition $k$ would be:

\begin{equation}
  c_k = P\left(\hat{x}_k = t_k | x = \mathbf{t}  \odot \mathbf{1}_{\mathcal{M}_l^{C}} \right)
\end{equation}

with $k=0,\dots, |\mathcal{T}| - 1$, $l=0,\dots, |\mathcal{S}| - 1$, and $\mathcal{M}_l$ being the set of masked token indexes for the state $S_l$.

For our Viterbi implementation, we want to maximize the product of the transition costs of the Viterbi path. Our approach can be summarized as follows:

\begin{enumerate}
  \item First, we define our state set $\mathcal{S}$ by masking the observed sequence $\mathbf{t}$ with all different $\mathcal{M}_l$ masked index combinations. We also initialized our cumulative cost to 1, and a new variable was used to store the optimal Viterbi path.
  \item For each state $\mathbf{s}_l$, we compute the output of the language model $\mathbf{y}_l=\mathrm{\textsc{lm}}
  (\mathbf{s}_l)$.
  \item We calculate the cost for each masked element $k$ in $\mathcal{M}_l$ by multiplying the corresponding index of the language model's output $\mathbf{y}_l(t_k)$ by the cumulative cost of the previous state. The previous state is easily derived by removing $k$ from the current state's masked index set $\mathcal{M}_l$.
  \item After getting each element's cost, we actualize our cumulative cost with the state maximum and add the element corresponding to the maximum argument to the optimal path.
  \item Once we have gone through all states, we return the optimal Viterbi path and cumulative cost.
\end{enumerate}

\begin{figure*}[!t]
\centering
\includegraphics[width=0.7\textwidth]{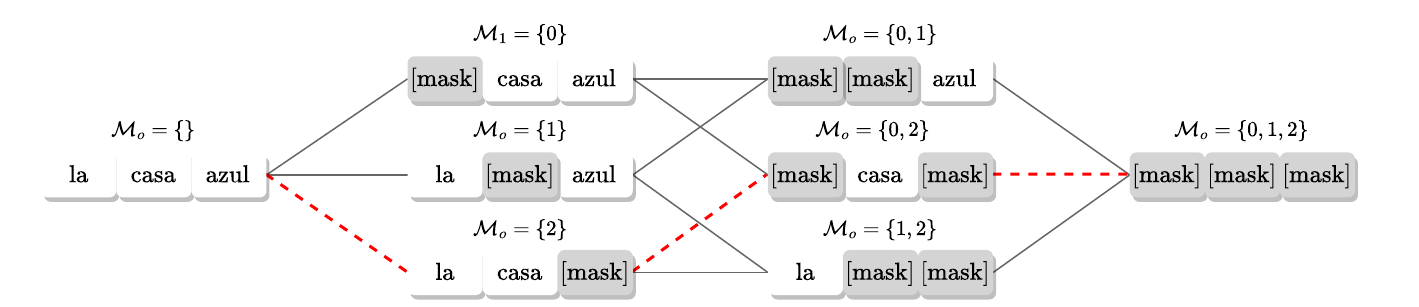}
\caption{Viterbi for maximum likelihood word ordering estimation, toy example with $N=3$. The optimal Viterbi path is in red.}
\label{fig:toy}
\end{figure*}

Figure \ref{fig:toy} shows a toy example with the phrase ``\textit{la casa azul}'' (``the blue house'').
In this implementation, our first state $S_0$ is the observed sequence $\mathbf{t}$ instead of a complete masked token sequence. This means that, for the proposed example, the optimal generation order would be \textit{casa} $\rightarrow$ \textit{la} $\rightarrow$ \textit{azul}.

\subsection{Spearman's rank correlation coefficient}
Once we find the maximum likelihood generation order, we need a metric to compare its differences from a causal generation order. 
For this, we need to measure the distance between our Viterbi state sequence $\mathcal{X}$ and the causal generation order. Several metrics in the literature measure the relationship between two sequences. Cosine similarity \citep{Srivastava2023} and Pearson's correlation coefficient \citep{Joost2016} are the most often used.

Nevertheless, our experiments are distinctive in that we compare two sequences that differ in order. For this reason, we consider Spearman's rank correlation coefficient \citep{Spearman1904} an appropriate evaluation metric.

Spearman's rank correlation coefficient, often denoted as $\rho$, measures how close to a monotonic function is the relationship between two variables. It is equivalent to the Pearson correlation of the ranks of two variables and is formulated as follows:
\begin{equation}
\rho = 1-\frac{6\sum_{i=0}^{N-1}d_i^2}{N\left(N^2 -1\right)}
\end{equation}
$N$ is the number of observations, and $d_i$ is the rank difference between the $i$-th observations of both variables.

\section{Results}
\label{sec:results}
\subsection{Language models}
\label{sec:model_results}
In this work, we used a Spanish implementation of GPT-2 base\footnote{Available at: \texttt{\url{https://huggingface.com/PlanTL-GOB-ES/gpt2-base-bne}}, October 2024} for causal language modeling and a Spanish implementation of RoBERTa base\footnote{Available at: \texttt{\url{https://huggingface.co/PlanTL-GOB-ES/roberta-base-bne}}, October 2024} for non-causal language modeling. Both models belong to MarIA \citep{gutierrezfandino2022}, the most relevant family of monolingual Spanish language models. However, the key ideas are transferable to other transformer language models with the same base architecture. Not only do these models have similar dimensions (12-layer, 768-hidden, 12-heads, 125M parameters of RoBERTa base vs. 12-layer, 768-hidden, 12-heads, 117M parameters of GPT-2 base), but they have also been trained on the same dataset and share the same tokenizer.

We also provide alternative sentence probability results using GPT-3 davinci-002\footnote{Available at: \texttt{\url{https://platform.openai.com/docs/models/gpt-base}}, October 2024}. The prompt used for the estimation was \textit{``Complete this sentence. You are acting as auto-complete. Simply complete the sentence to the best of your ability, make sure it is just one word''}. Then, individual generation probabilities were calculated through the logprobs obtained using the OpenAI API.

{Even though it is out of the main focus of this paper, for the benefit of the interested reader, we performed a preliminary experiment in English, using OpenAI's GPT-2 base\footnote{Available at: \texttt{\url{https://huggingface.com/gpt2}}, October 2024} for non-causal language modeling and Google's BERT\footnote{Available at: \texttt{\url{https://huggingface.co/bert-base-cased}}, October 2024} for causal language modeling.

\subsection{Data set}
\label{sec:data_results}
Our experiments use a data set composed of 7,044 short Spanish sentences, of which 3,288 are declarative, and 3,756 are interrogative. These sentences are drawn from all possible permutations of 1,174 different subject-verb-object triplets, 548 for declarative sentences and 626 for interrogative sentences. All these triplets were manually crawled from diverse educational resources and constituted entirely grammatically correct sentences in all six possible orderings. All sentences are coherent, adequately capitalized, and punctuated.

For the preliminary experiments in English, we took a subset of 120 declarative sentences in this language, all of them with \textsc{svo} structure, as this is the only grammatically correct syntax for English declarative sentences.

All data sets are available in a Zenodo repository\footnote{Available at \url{https://doi.org/10.5281/zenodo.13936987}, October 2024}. The code used is available upon reasonable request to the corresponding author.

\subsection{Results and discussion}
In our experiments, we have computed the maximum likelihood generation order for each of the sentences in the data set in Section \ref{sec:data_results} using the Viterbi algorithm implementation described in Section \ref{sec:method}.

\begin{table}[!t]
\caption{Average generation probability per syntactic structure from declarative sentences}
\label{tab:declarative_results}
\centering
\begin{tabular}[t]{l c c c}\toprule
\textbf{Structure}& \textbf{RoBERTa} & \textbf{GPT-2} & \textbf{GPT-3}\\
\midrule
\textsc{svo} & $\mathbf{1.596 \times 10^{-9}}$ & $\mathbf{2.428 \times 10^{-10}}$ & $\mathbf{6.858 \times 10^{-18}}$\\
\textsc{sov} & \num{1.301e-12} & 
$\mathbf{2.268 \times 10^{-15}}$ & $\mathbf{7.846 \times 10^{-25}}$\\
\textsc{vso} & \num{1.257e-11} & \num{3.266e-13} & \num{7.798e-21}\\
\textsc{vos} & \num{3.148e-11} &  \num{1.783e-13} & \num{3.413e-22}\\
\textsc{osv} & \num{2.011e-13} &  $\mathbf{6.711 \times 10^{-16}}$ & $\mathbf{5.682 \times 10^{-27}}$\\
\textsc{ovs} & \num{3.976e-12} &  \num{5.065e-14} & \num{5.043e-22}\\
\bottomrule
\end{tabular}
\end{table}

\begin{figure}
\centering
\includegraphics[width=0.7\linewidth]{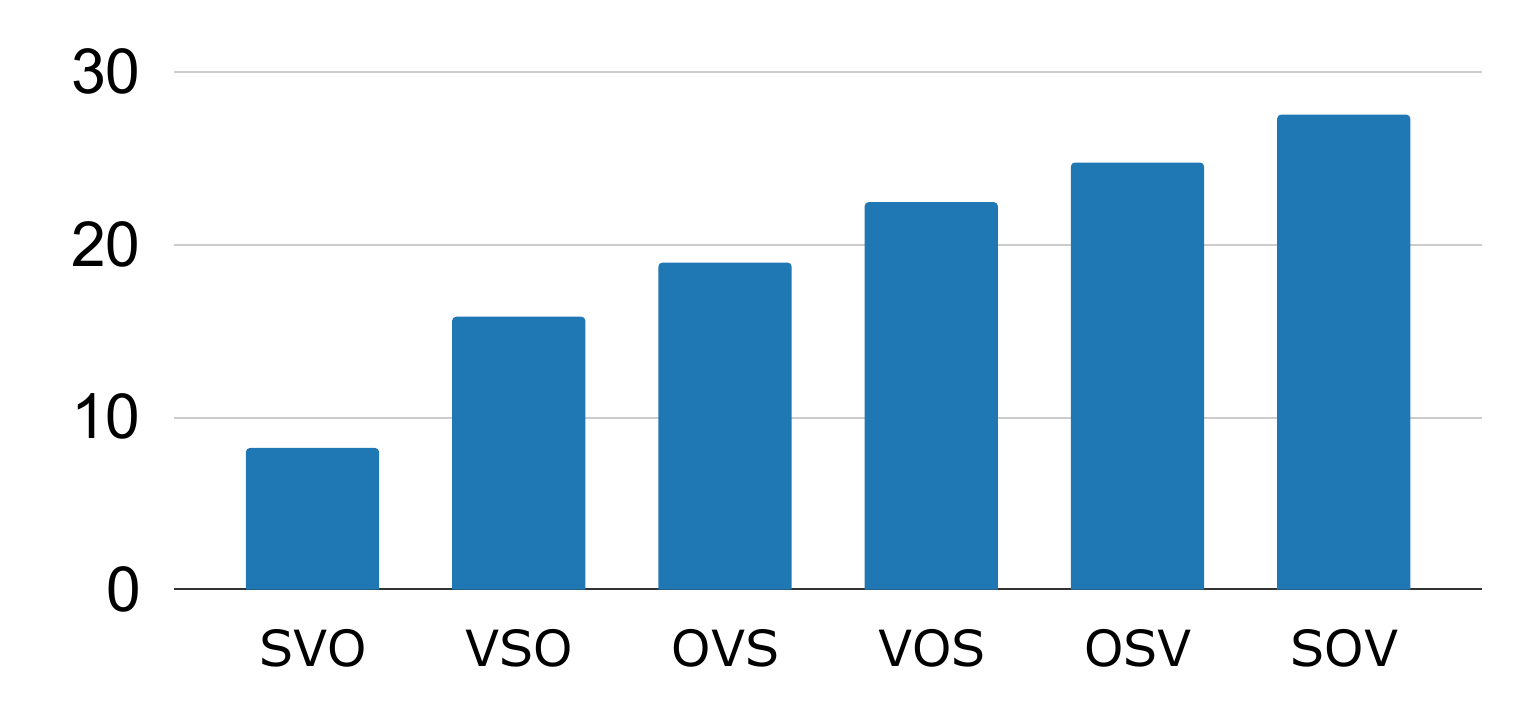}
\caption{Optimal-causal generation probability ratio in logarithmic units from declarative sentences}
\label{fig:log_results_dec}
\end{figure}

\begin{table}[!t]
\caption{Average generation probability per syntactic structure for interrogative sentences}
\label{tab:interrogative_results}
\centering
\begin{tabular}[t]{l c c c}\toprule
\textbf{Structure}& \textbf{RoBERTa} & \textbf{GPT-2} & \textbf{GPT-3}\\
\midrule
\textsc{svo} &\num{5.955e-12} &  $\mathbf{1.616 \times 10^{-12}}$ & $\mathbf{9.257 \times 10^{-24}}$\\
\textsc{sov} & \num{1.343e-14} & $\mathbf{3.206 \times 10^{-17}}$ & $\mathbf{1.732\times 10^{-33}}$\\
\textsc{vso} & \num{1.219e-12} & \num{1.878e-13} & \num{3.844e-26}\\
\textsc{vos} & $\mathbf{1.773 \times 10^{-12}}$ & \num{3.140e-14} & \num{5.470e-27}\\
\textsc{osv} & \num{1.521e-15} &  $\mathbf{2.155 \times 10^{-18}}$ & $\mathbf{3.839 \times 10^{-29}}$\\
\textsc{ovs} & \num{2.507e-14} &  \num{7.954e-17} & \num{7.979e-26}\\
\bottomrule
\end{tabular}
\end{table}

\begin{figure}
\centering
\includegraphics[width=0.7\linewidth]{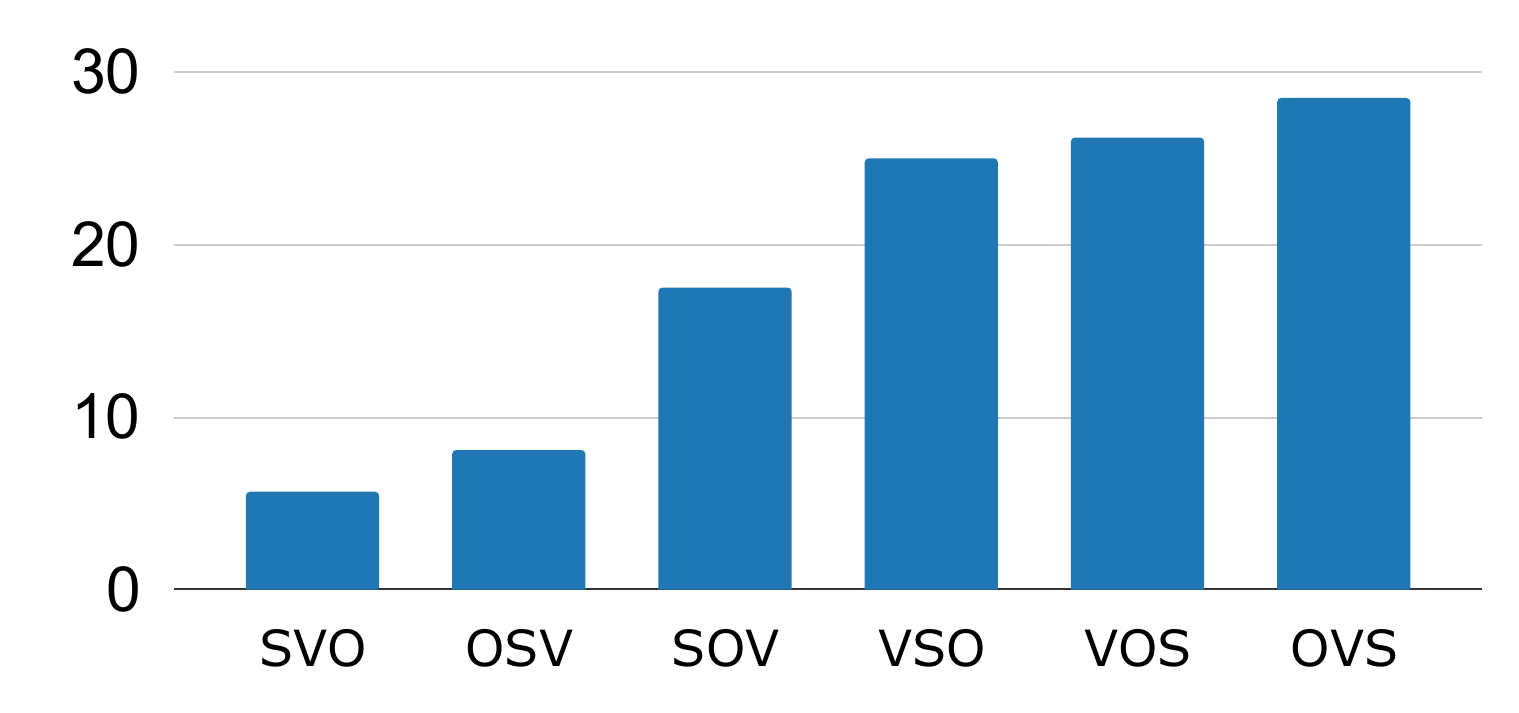}
\caption{Optimal-causal generation probability ratio in logarithmic units for interrogative sentences}
\label{fig:log_results_int}
\end{figure}

Table \ref{tab:declarative_results} shows the average sequence generation probability for declarative sequences using our maximum likelihood generation order with the non-causal language model RoBERTa, compared to the same metric for the causal GPT-2 and GPT-3 models. As expected, the maximum likelihood non-causal approach leads to higher generation probability for all grammatical structures. However, this difference is narrower for sentences with a \textsc{svo} structure. This is even more notable in Figure \ref{fig:log_results_dec}, which shows the results of $10\log_{10}(\frac{P_{\mathrm{roBERTa}}}{P_{\mathrm{GPT-2}}})$ for all possible syntactic structures within declarative sentences. In contrast, GPT-2 seems to avoid declarative sequences ending with a verb. This may be attributed to the pivotal function of the verb in establishing a connection between the subject and the object, hence making predicting the object based on the subject, and vice versa, more challenging in the absence of the verb.

\begin{table}[!t]
\caption{Average generation probability for English declarative sentences}
\label{tab:english_results}
\centering
\begin{tabular}[t]{c c c}\toprule
 \textbf{BERT} & \textbf{GPT-2} & \textbf{GPT-3}\\
\midrule
\num{3.199e-10} &  \num{4.901e-10} & \num{7.761e-12}\\
\bottomrule
\end{tabular}
\end{table}
As previously stated in Section \ref{sec:model_results}, RoBERTa and GPT-2 models have similar parameters and were trained on the same data. Therefore, the differences are likely due to generation order. The order is still highly variable even if the \textsc{svo} structure is also frequent in Spanish, thus; such generation order bias may hinder the intrinsic richness of the Spanish language. This particularity of the Spanish language, and potentially other Romance languages, is further substantiated by the preliminary results observed with the English data set. In this case, Table \ref{tab:english_results} shows that the generation probabilities of both causal and non-causal language models are very similar, with the additional insight that  GPT-3 and GPT-2 results are closer than the Spanish case. This may result from the predominance of English in the training data of GPT-3, which significantly impacts both the generation probability and the number of tokens per word in both languages.

Table \ref{tab:interrogative_results} shows the generation probabilities by syntactic structure for interrogative sentences. For causal ordering, the \textsc{svo} structure has the highest generation probability with causal language models GPT-2 and GPT-3, whereas, as in the case of declarative sentences,  \textsc{sov} and \textsc{osv} have the lowest generation probability. However, the structure with the highest generation probability for non-cause optimal word generation order is \textsc{vos}. As we can see in Figure \ref{fig:log_results_int}, for interrogative sentences, the most significant difference between causal and optimal non-causal generation probability occurs in the sentences ending with the subject.

Focusing on optimal generation order instead of generation probability, Figure \ref{fig:rho_all} shows the histogram of Spearman's rank correlation coefficient $\rho$ of causal generation order compared to our Viterbi maximum likelihood generation order for all sentences. As we can see, the maximum likelihood generation order shows a solid positive monotonic relationship with causal generation order (i.e., $\rho > 0.8$) for less than 5\% of the declarative sequences. Furthermore, since the majority of values lie on the negative axis of the histogram, our results reveal that optimal word generation order tends to be closer to right-to-left than to left-to-right text generation.

In Figure \ref{fig:rho_structures}, we see these results through syntactic structure. Even if optimal generation order is never closely aligned with causality, some remarkable patterns are apparent. One of the most striking is for the \textsc{svo} structure, as more than $25\%$ of the analyzed sequences had a $\rho$ on the $\left[-0.3189 -0.2471\right)$ interval. Of the 139 samples within this interval, 83 followed an object-subject-verb generation order. Most exceptions corresponded to either a single-word subject, which is most likely to be generated first due to capitalization, or to attributive sentences.

We observe similar phenomena for the \textsc{sov} structure, with a peak in $\rho = \left[-0.1451, -0.0674\right)$. Of the 98 sentences in this interval, 61 followed an object-subject-verb generation order, with most exceptions being analogous to those in the \textsc{svo} case.

There is also a peak in the $\rho = \left[-0.2031, -0.1383\right)$ interval in the \textsc{vso} histogram. In 60 of the 72 samples in this interval, the subject was the last element of the optimal generation order.

\begin{figure}
\centering
\includegraphics[width=0.52\linewidth]{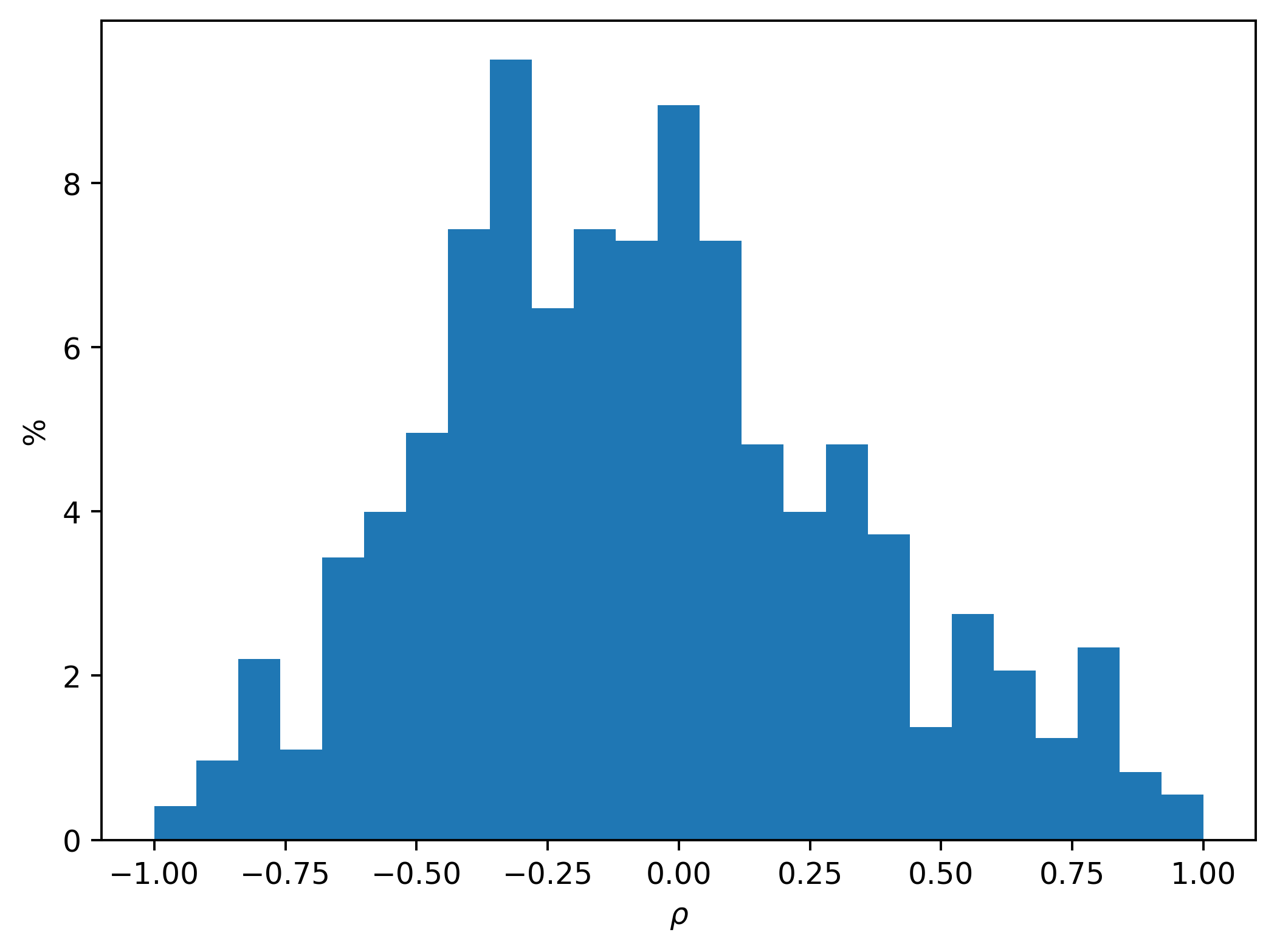}
\caption{Histogram of Spearman's rank correlation coefficient ($\rho$) for declarative sentences.}
\label{fig:rho_all_int}
\end{figure}

\begin{figure*}[!t]
\centering
\subfloat[]{\includegraphics[width=0.25\textwidth]{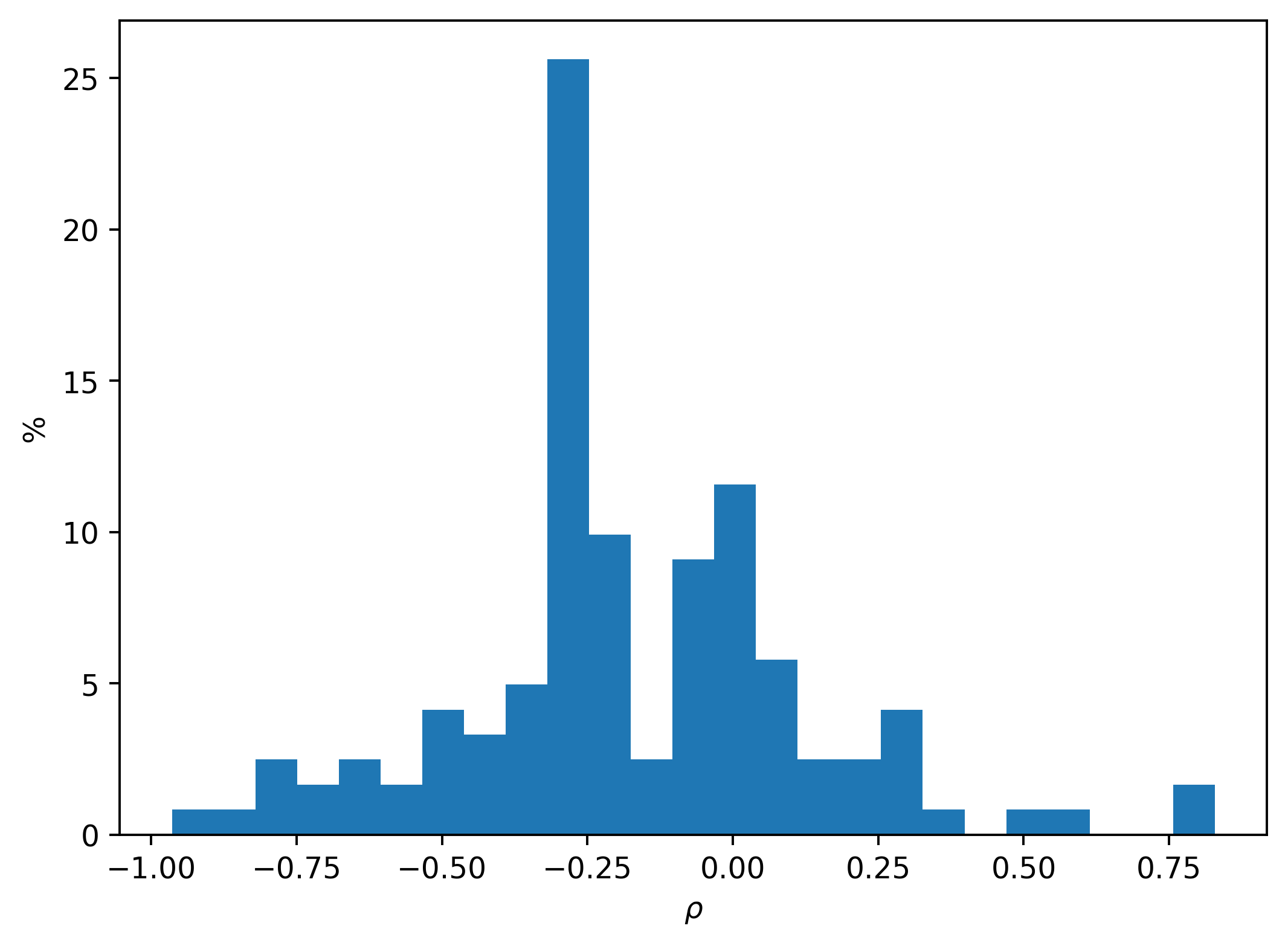}
}
\hfil
\subfloat[]{\includegraphics[width=0.25\textwidth]{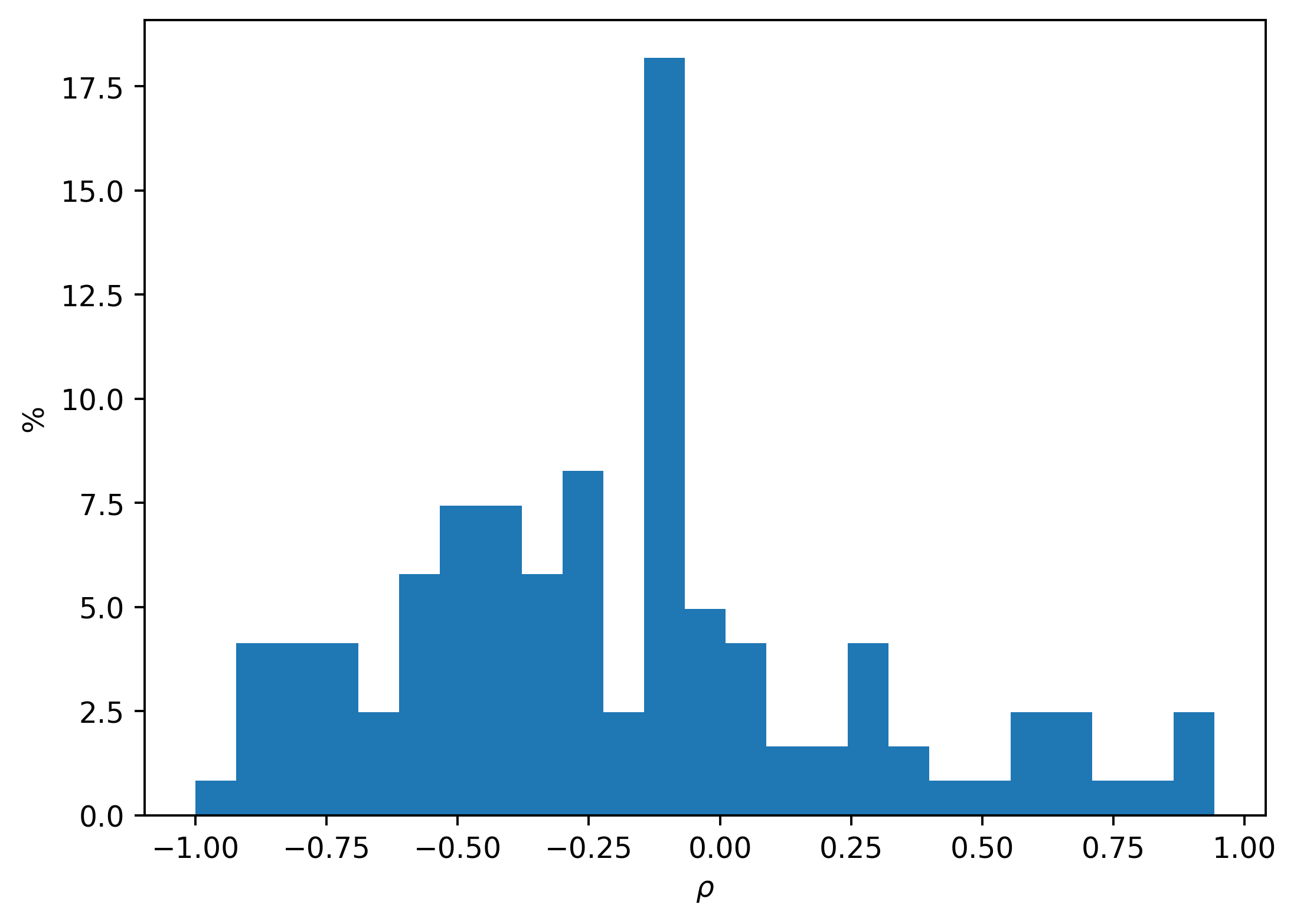} 
}\\
\subfloat[]{\includegraphics[width=0.25\textwidth]{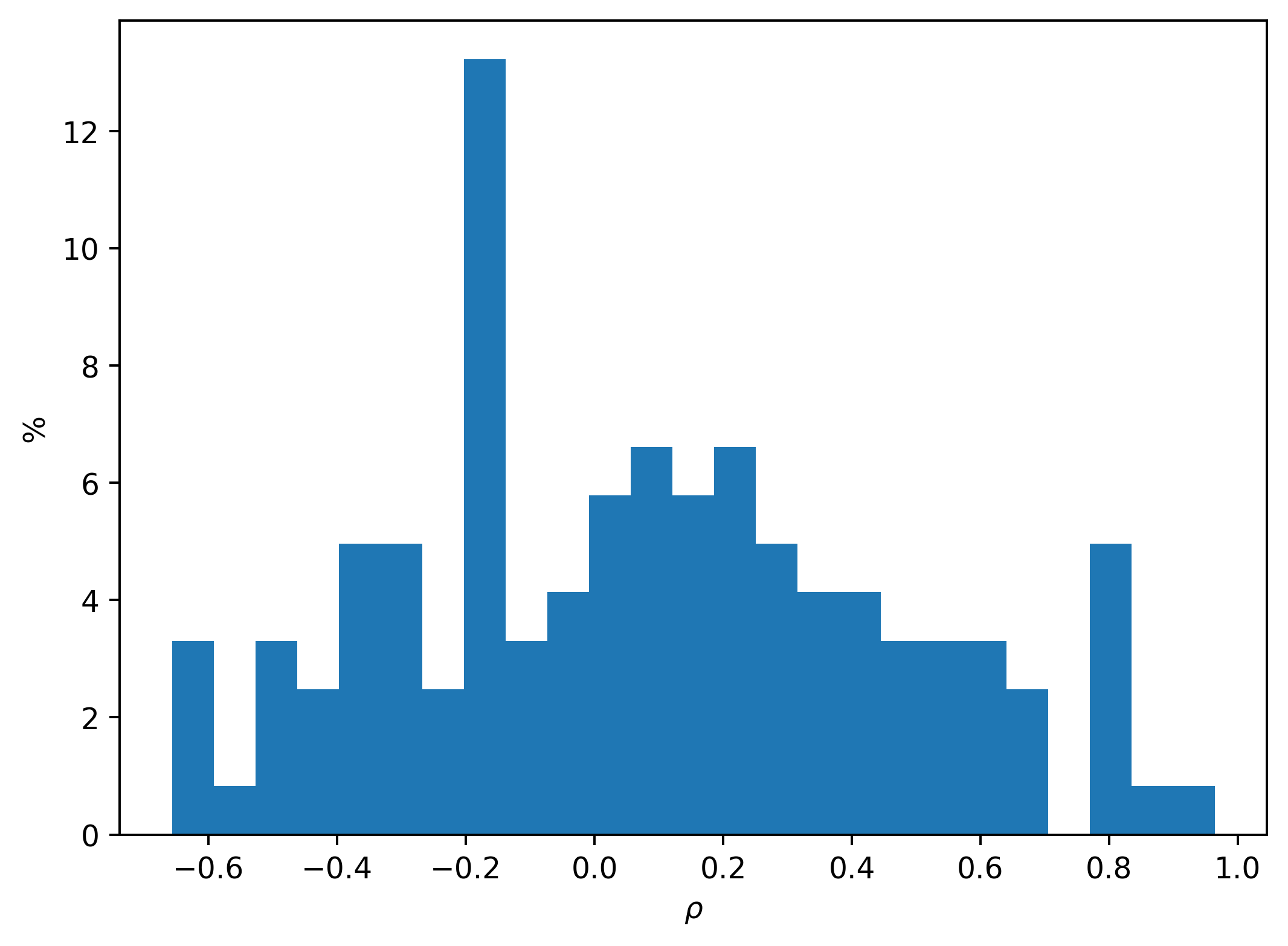}
}
\hfil
\subfloat[]{\includegraphics[width=0.25\textwidth]{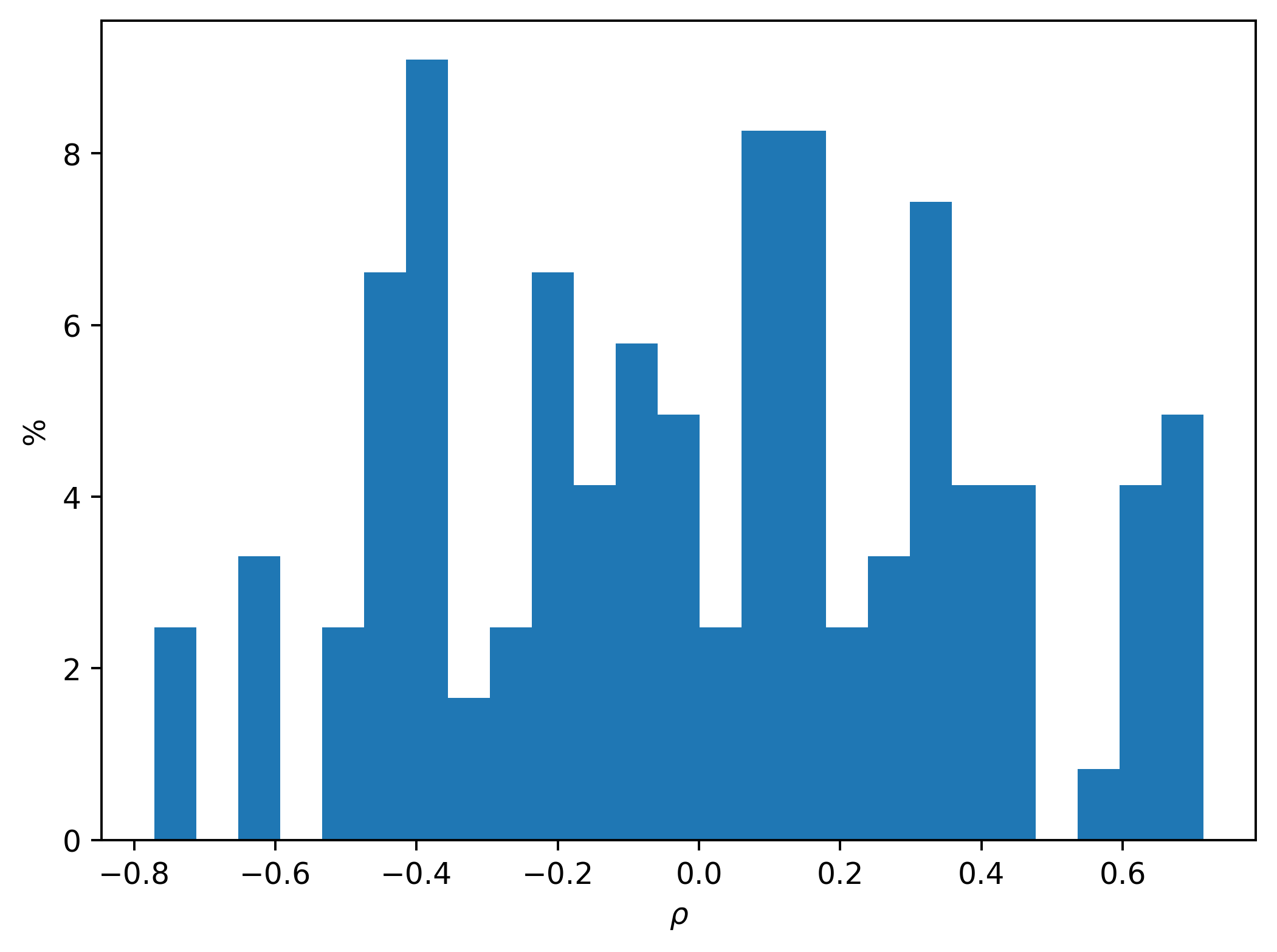}
}\\
\subfloat[]{\includegraphics[width=0.25\textwidth]{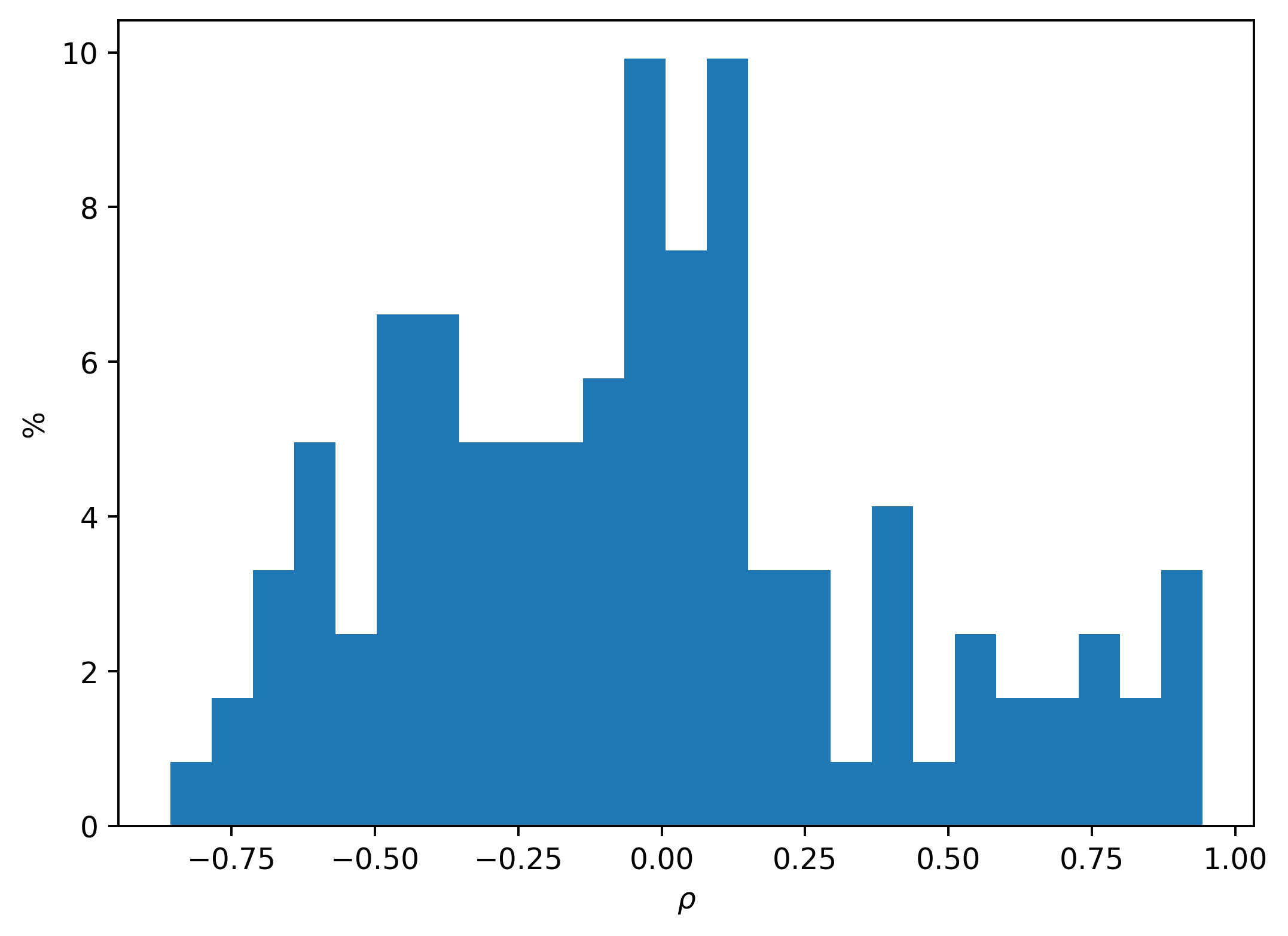}
}
\hfil
\subfloat[]{\includegraphics[width=0.25\textwidth]{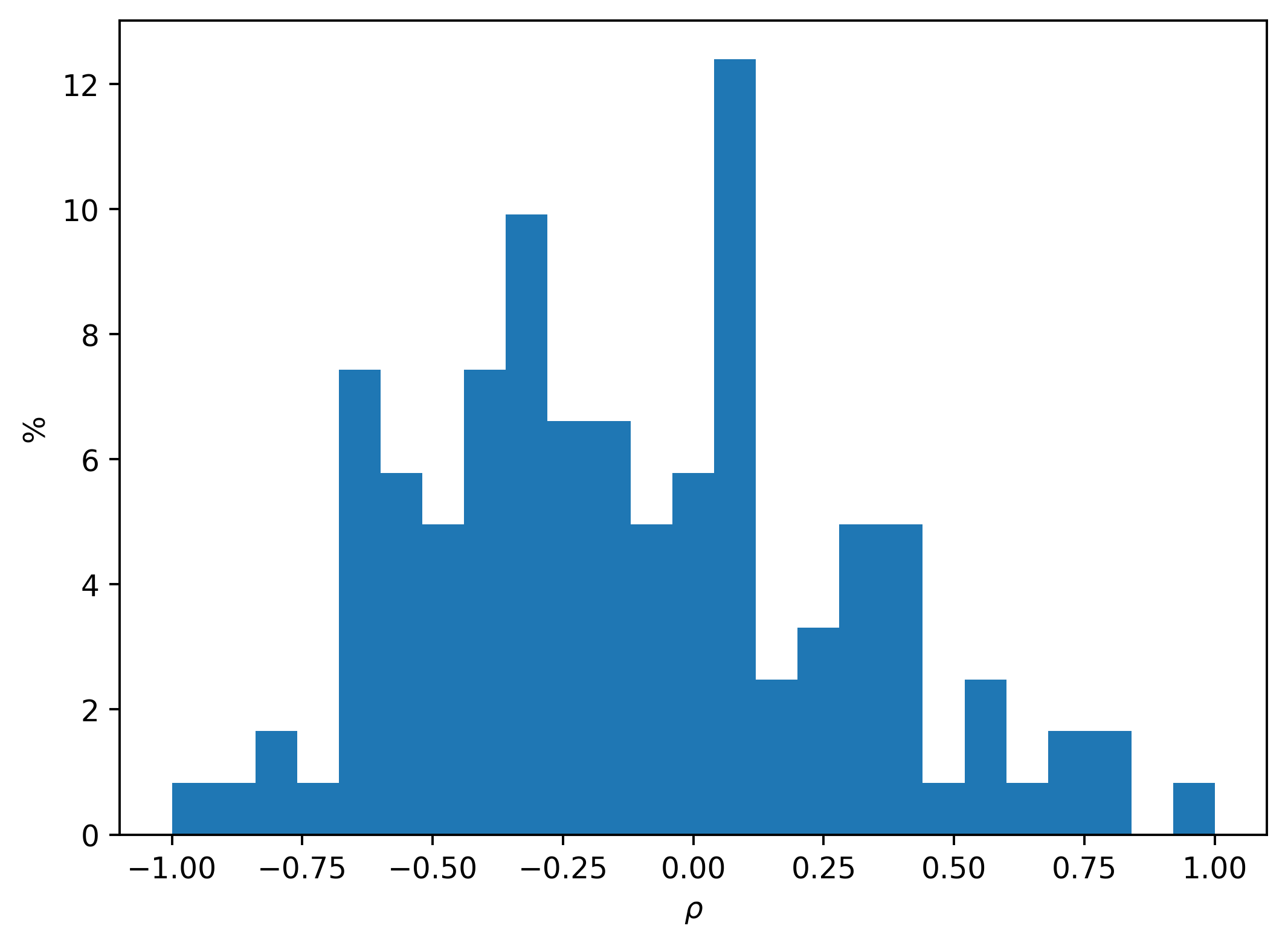}
}
\caption{Histogram of Spearman's rank correlation coefficient ($\rho$) by syntactic structure for declarative sentences. (a) \textsc{svo}. (b) \textsc{sov}. (c) \textsc{vso}. (d) \textsc{vos}. (e) \textsc{osv}. (f) \textsc{ovs}.}
\label{fig:rho_structures}
\end{figure*}

As shown in Figure \ref{fig:rho_all_int}, interrogative sentences' optimal non-causal order is closer to causal than in the case of declarative sentences. However, only the optimal word order of 6\% of the analyzed interrogative sentences was strongly aligned with causality. Figure \ref{fig:rho_structures_int} shows these results by syntactic structure. As in declarative sentences, we can see some patterns, among which the \textsc{osv} ordering stands out, with more than 17\% of the analyzed sequences within the $\rho = \left[0.4262, 0.4888\right)$ interval. Of the 108 sentences within this interval 104, the verb was the last element of the optimal generation order.

\begin{figure}
\centering
\includegraphics[width=0.52\linewidth]{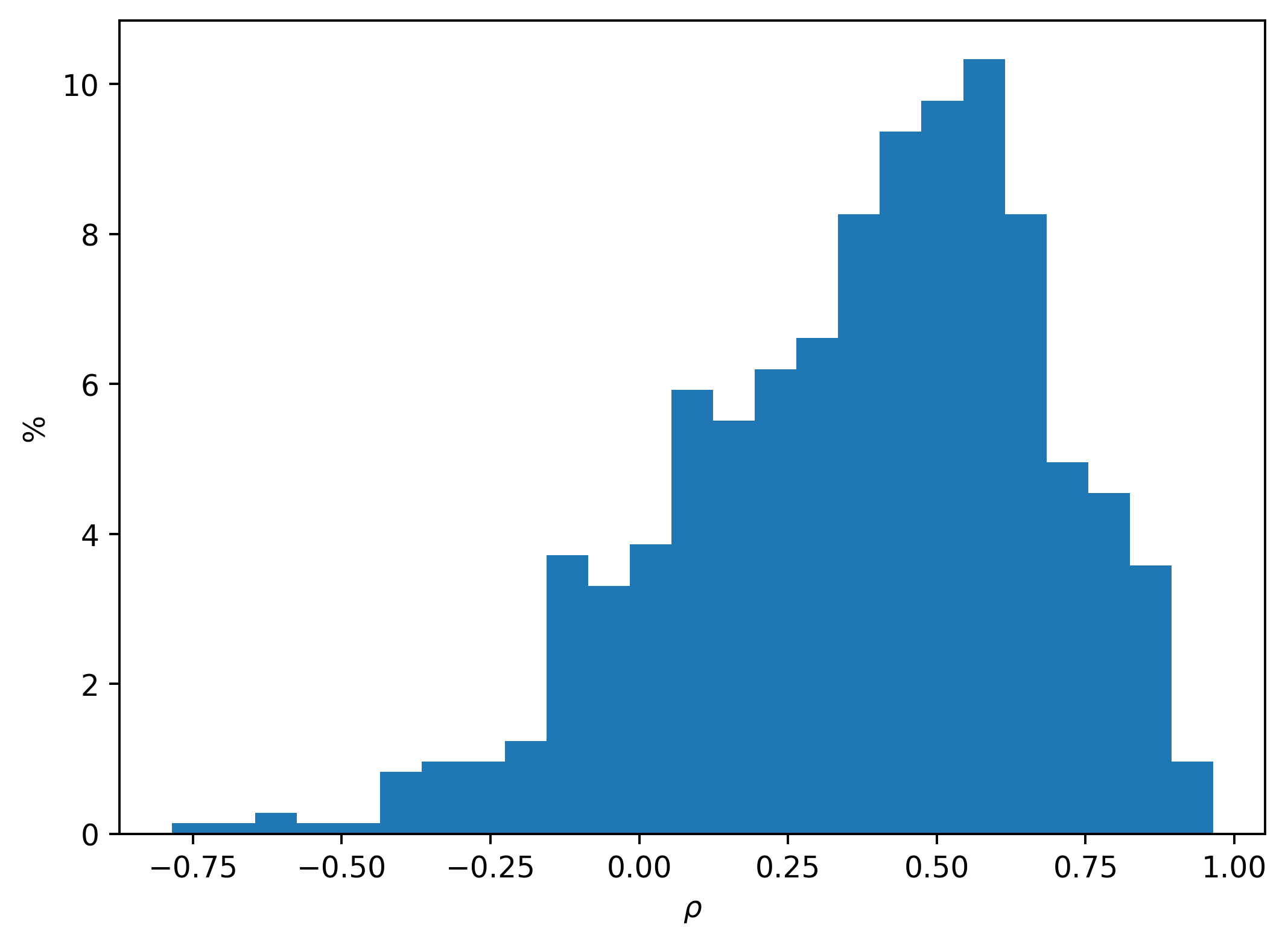}
\caption{Histogram of Spearman's rank correlation coefficient ($\rho$) for interrogative sentences.}
\label{fig:rho_all}
\end{figure}

\begin{figure*}[!t]
\centering
\subfloat[]{\includegraphics[width=0.22\textwidth]{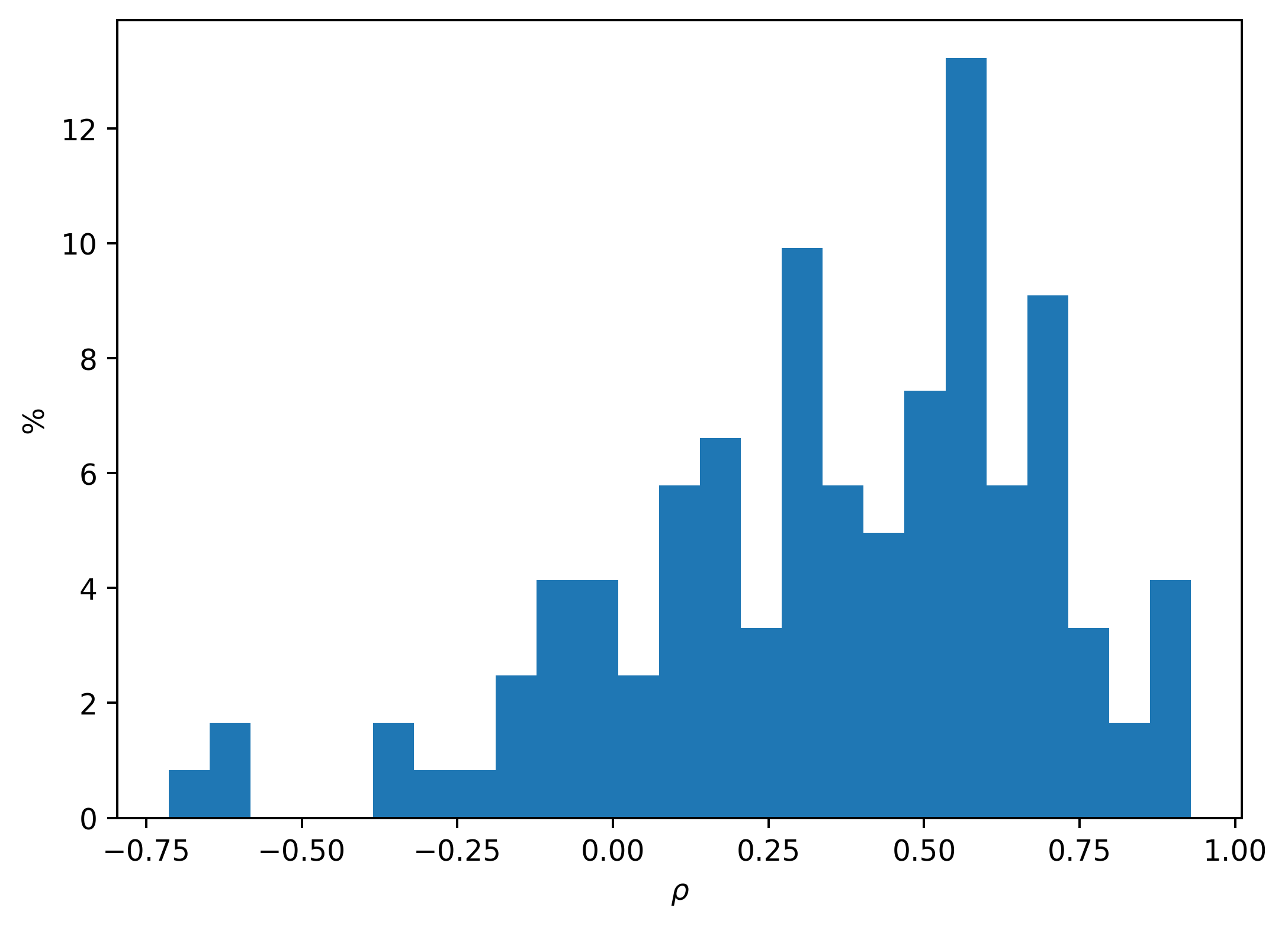}
}
\hfil
\subfloat[]{\includegraphics[width=0.22\textwidth]{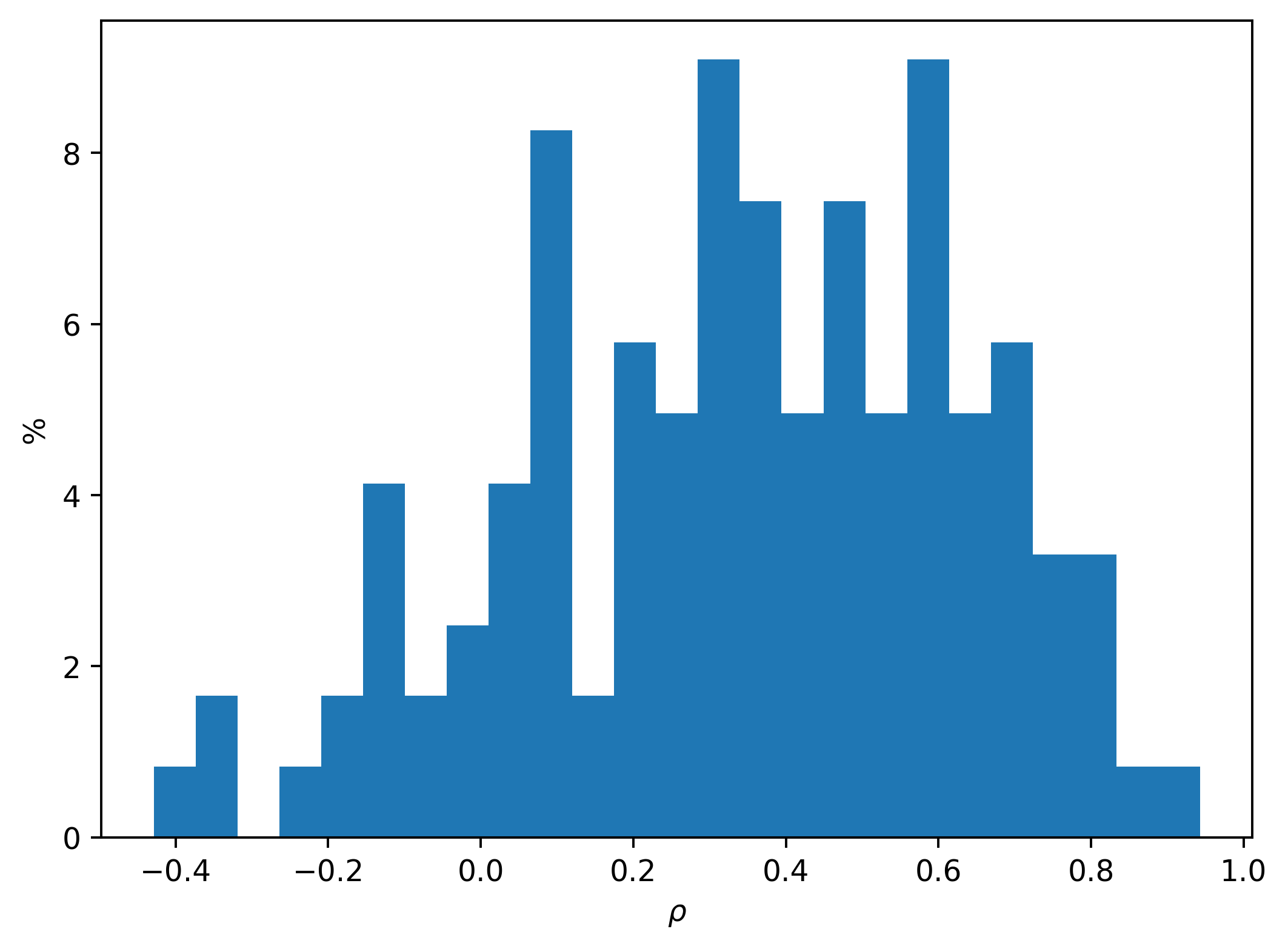} 
}\\
\subfloat[]{\includegraphics[width=0.22\textwidth]{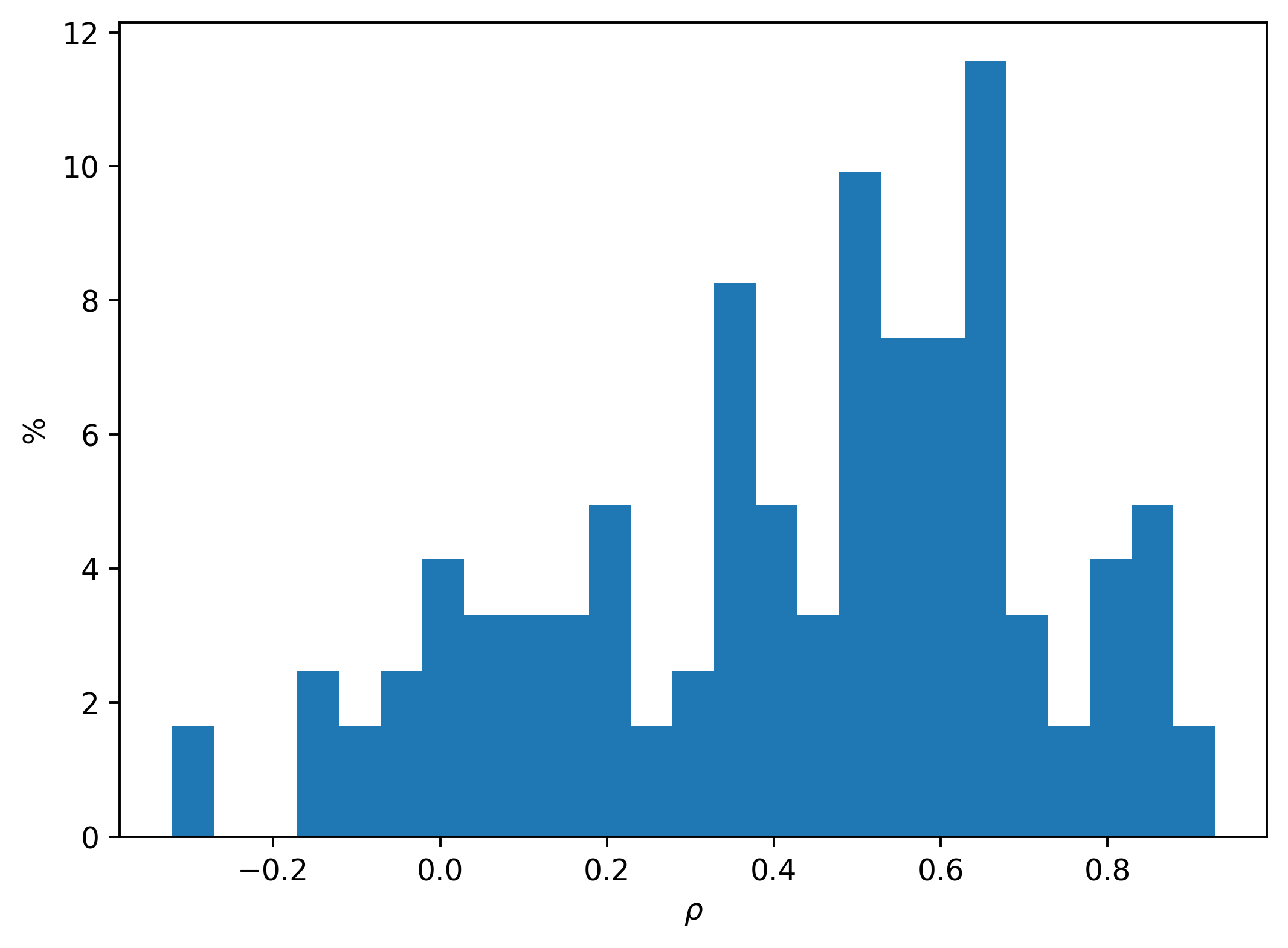}
}
\hfil
\subfloat[]{\includegraphics[width=0.22\textwidth]{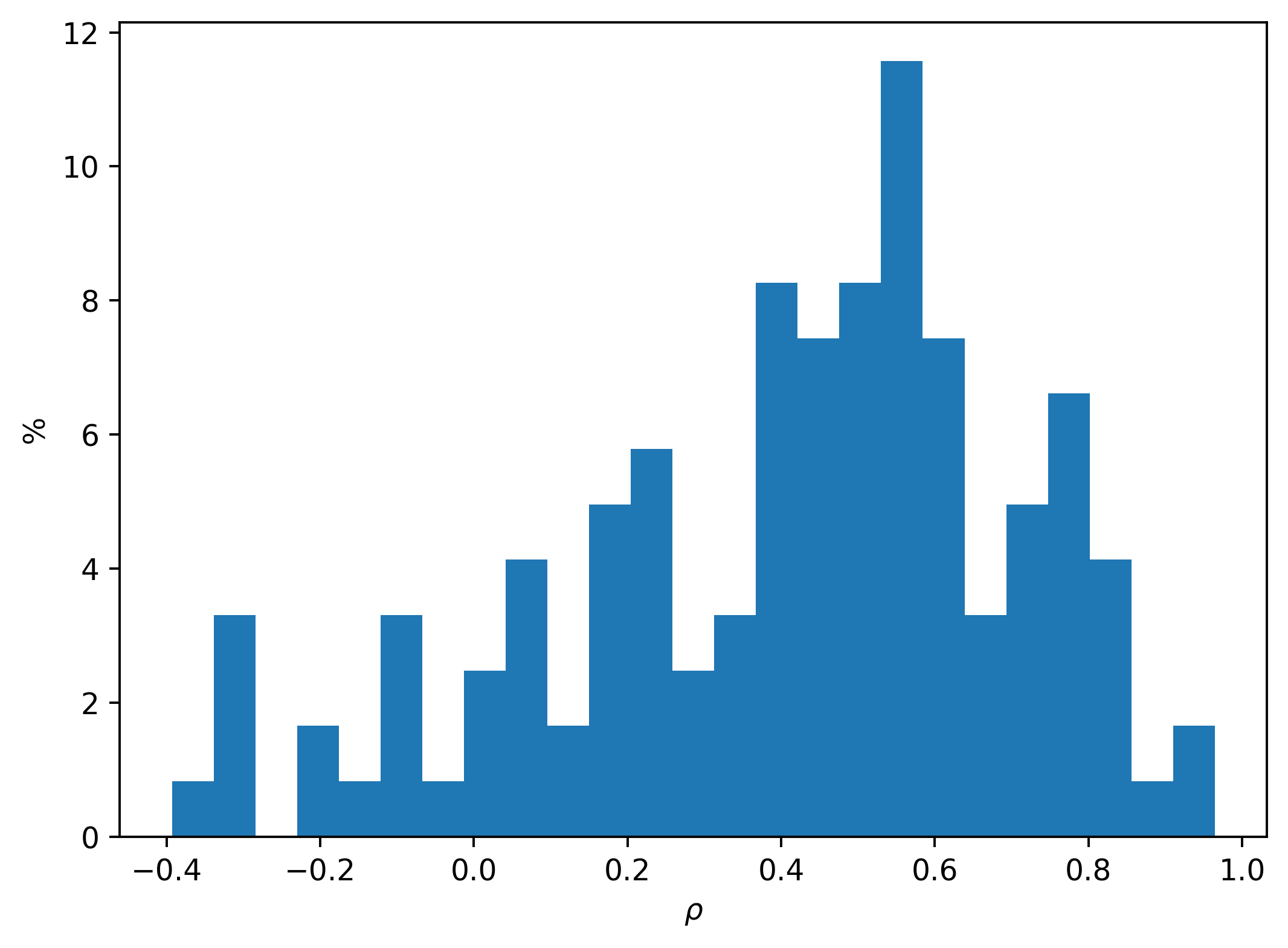}
}\\
\subfloat[]{\includegraphics[width=0.22\textwidth]{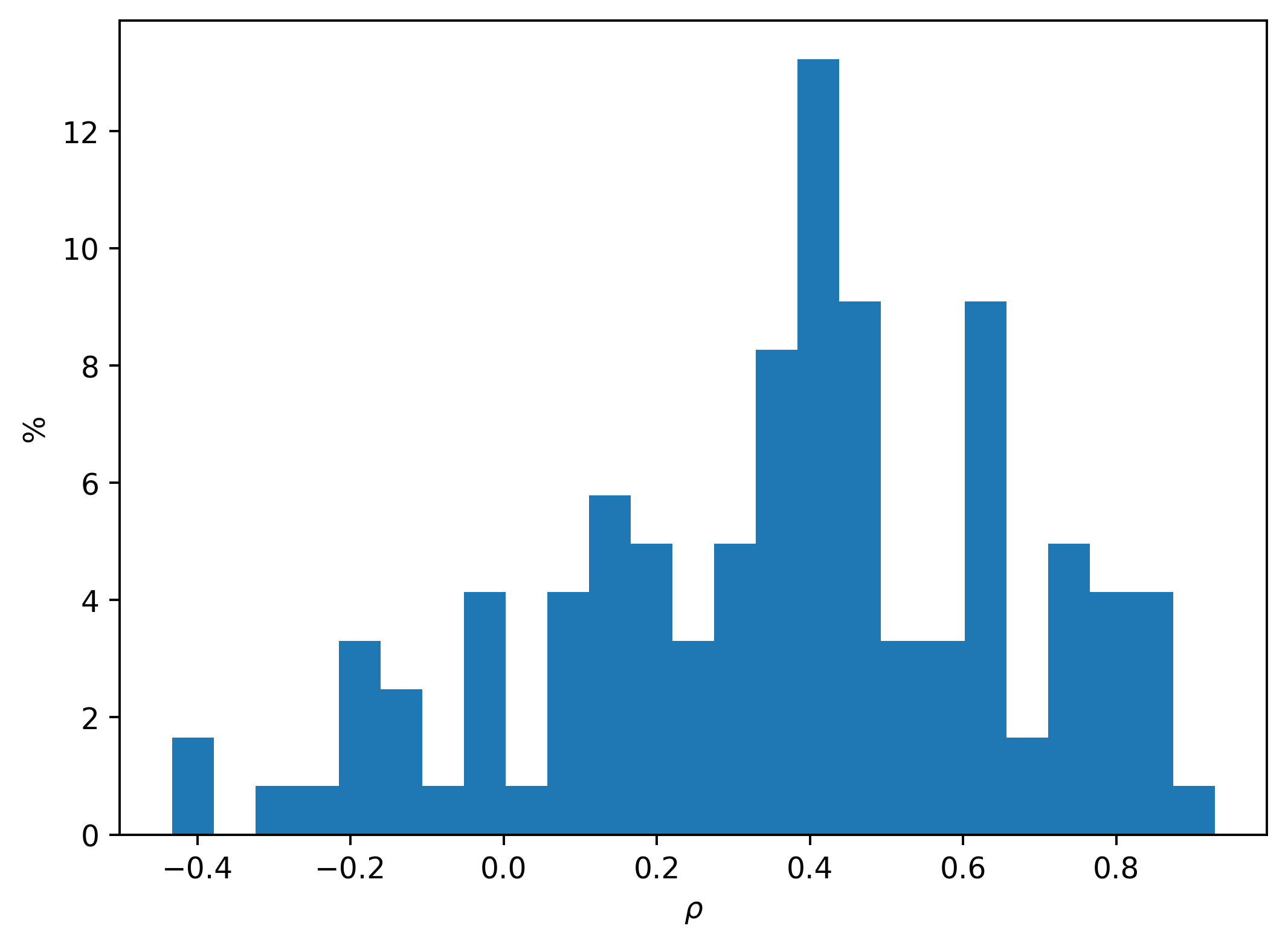}
}
\hfil
\subfloat[]{\includegraphics[width=0.22\textwidth]{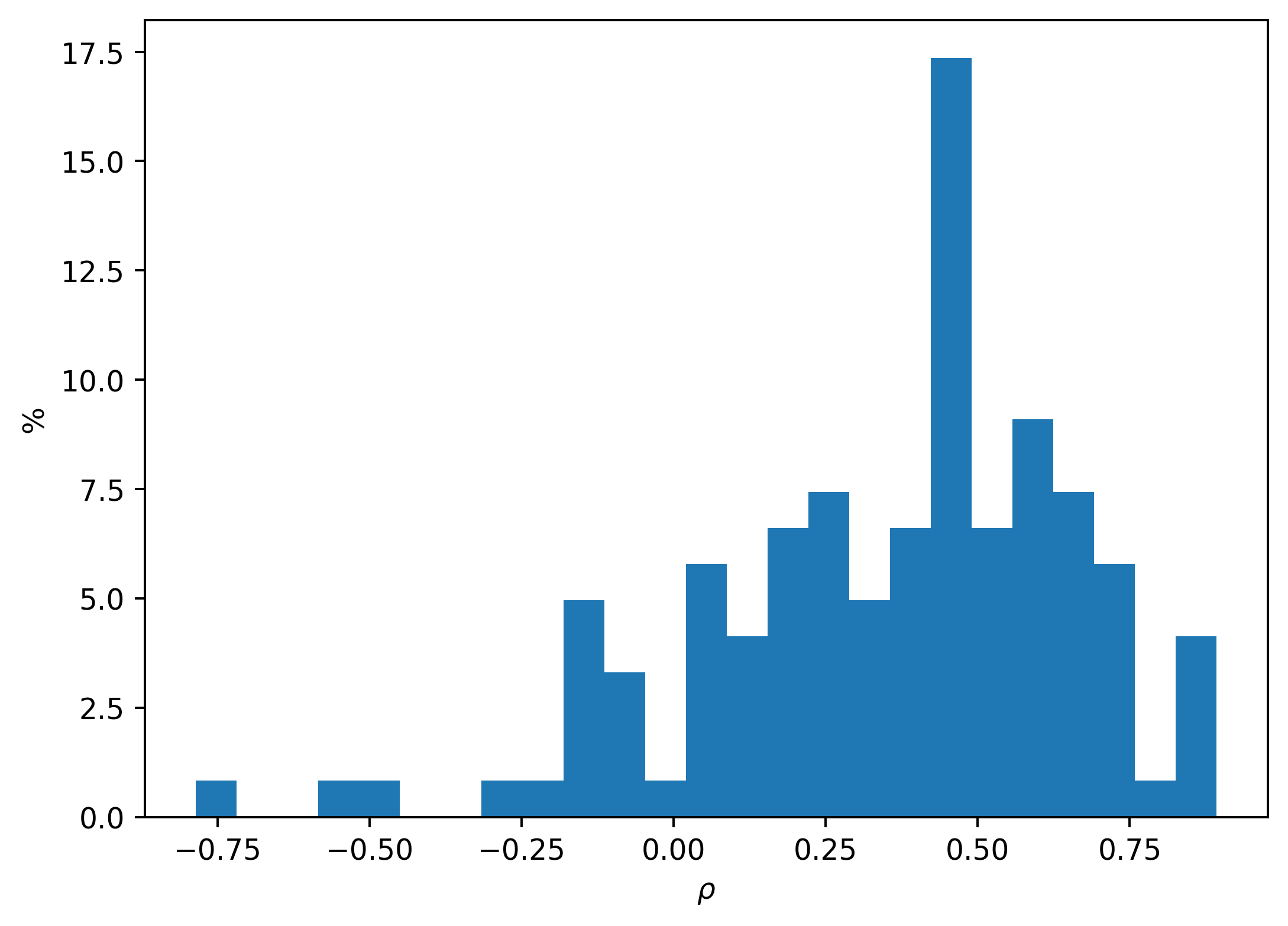}
}
\caption{Histogram of Spearman's rank correlation coefficient ($\rho$) by syntactic structure for interrogative sentences. (a) \textsc{svo}. (b) \textsc{sov}. (c) \textsc{vso}. (d) \textsc{vos}. (e) \textsc{osv}. (f) \textsc{ovs}.}
\label{fig:rho_structures_int}
\end{figure*}

\section{Conclusions}
\label{sec:conclusions}

In this paper, we have presented the first approach to optimal word order estimation for natural language generation using the Viterbi algorithm, particularizing it to the study of Spanish \textsc{nlg}. We have conducted a comparative experiment between the generation probability of our maximum likelihood estimate and that of the mainstream left-to-right \textsc{nlg} approach. Our findings suggest that causal \textsc{nlg} systems favor \textsc{svo}, restricting the intrinsic flexibility of the Spanish language. We have also analyzed the relation between the ideal (optimal) non-causal generation order and the causal ordering of decoder-only transformer language models using Spearman's rank correlation coefficient, proving that the ideal order predicted by the maximum likelihood estimator differs significantly from causal order and depends on the syntactic structure of the target sentence. In this regard, some Spearman's rank correlation histogram peaks suggest relevant patterns such as an object-subject-verb preference in the optimal generation order for \textsc{svo} and \textsc{sov} declarative sentences, which deserve future work.

In future research, we will conduct a more in-depth analysis of the maximum likelihood text generation order for more complex syntactic structures, including passive and impersonal phrases. We will also apply our findings to practical Spanish non-causal \textsc{nlg} by adding cost policy agents to the text generation process. Following the reasoning presented in this paper, in which we modeled word generation order as a Markov decision process, we plan to incorporate an agent to non-causal \textsc{nlg} to learn the optimal word generation order via reinforcement learning. We intend to explore deeper into the phenomena underlying the results of this experiment with more linguistics and psycholinguistics-oriented approaches.

We hope that our findings will contribute to the advancement of novel non-causal Spanish \textsc{nlg} approaches and encourage comparable investigations in languages other than English using bidirectional transformers.

\section*{Acknowledgements}

This work was partially supported by Xunta de Galicia grants ED481B-2021-118, ED481B-2022-093, and ED431C 2022/04, Spain; Ministerio de Educación grant FPU21/00798, Spain; and Ministerio de Ciencia e Innovación grant TED2021-130824B-C21, Spain.

\bibliographystyle{model5-names}
\bibliography{refs}

\end{document}